\documentclass[11pt]{article}
\pdfoutput=1
\usepackage{acl}
\usepackage{graphicx}
\usepackage{multirow}
\usepackage{multicol}
\usepackage{makecell}
\usepackage{times}
\usepackage{latexsym}
\usepackage{todonotes}
\usepackage{subcaption}
\usepackage{expex}
\usepackage[T1]{fontenc}

\usepackage[utf8]{inputenc}

\usepackage{microtype}

\title{XNLIeu: a dataset for cross-lingual NLI in Basque}

\author{Maite Heredia$^{1}$ \quad Julen Etxaniz$^{1}$ \quad Muitze Zulaika$^{2}$ \\
  \bf Xabier Saralegi$^{2}$ \quad Jeremy Barnes$^{1}$ \quad Aitor Soroa$^{1}$ \\
  $^{1}$HiTZ Center - Ixa, University of the Basque Country UPV/EHU \\
  $^{2}$Orai NLP Technologies\\
\texttt{\{maite.heredia\}@ehu.eus} \\
}

\begin{document}
\maketitle
\begin{abstract}
XNLI is a popular Natural Language Inference (NLI) benchmark widely used to evaluate cross-lingual Natural Language Understanding (NLU) capabilities across languages. In this paper, we expand XNLI to include Basque, a low-resource language that can greatly benefit from transfer-learning approaches. The new dataset, dubbed \textrm{XNLIeu}, has been developed by first machine-translating the English XNLI corpus into Basque, followed by a manual post-edition step. We have conducted a series of experiments using mono- and multilingual LLMs to assess a) the effect of professional post-edition on the MT system; b) the best cross-lingual strategy for NLI in Basque; and c) whether the choice of the best cross-lingual strategy is influenced by the fact that the dataset is built by translation. The results show that post-edition is necessary and that the translate-train cross-lingual strategy obtains better results overall, although the gain is lower when tested in a dataset that has been built natively from scratch. Our code and datasets are publicly available under open licenses\footnote{\url{https://github.com/hitz-zentroa/xnli-eu}}.
\end{abstract}

\section{Introduction}
\label{sec:intro}
The Natural Language Inference (NLI) task consists in classifying pairs of sentences --a premise and a hypothesis-- according to their semantic relation: \textit{entailment}, when the meaning of the premise entails that of the hypothesis; \textit{contradiction}, when both sentences have opposing truth conditions and can not co-occur at the same time; and \textit{neutral}, when both sentences are not semantically related (see Table \ref{tab:examples} for examples).

NLI is an important task towards Natural Language Understanding (NLU), and is often used to test the semantic understanding of language models. It provides a general framework where different NLP tasks can be reframed, including information retrieval \citep{retrieval-dusek2023improving}, metaphor detection \citep{figurative-stowe-etal-2022-impli} or relation extraction \citep{relation-sainz-etal-2021-label}. The NLI paradigm has also been proposed as a way to detect hallucination in Natural Language Generation (NLG) ~\citep{10.1145/3571730}.

\begin{table}[t]
\centering
\resizebox{\columnwidth}{!}{%
\begin{tabular}{ll}
\hline
\textbf{premise}       & Yesterday I saw an octopus at the beach. \\ \hline
\textbf{entailment}    & I was at the beach yesterday.            \\
\textbf{contradiction} & Yesterday I spent the whole day at home. \\
\textbf{neutral}       & Octopi are my favourite animals.      \\ \hline
\end{tabular}%
}
\caption{Example of a premise and three different hypotheses with the three possible relations.}
\label{tab:examples}
\end{table}

XNLI~\cite{xnli-conneau-etal-2018} is a popular benchmark widely used to evaluate cross-lingual NLI capabilities among languages. It comprises $7,500$ premise/hypothesis pairs in English that were manually translated to 14 high- and low-resource languages. In this paper we expand XNLI to include Basque, a low-resource language spoken in Spain and France (ISO-code: \textit{eu}). The new dataset, dubbed \textrm{XNLIeu}, has been built by machine translating and post-editing the English XNLI. We release both the post-edited and machine-translated versions, which we used to assess to what extent professional post-edition is necessary to obtain a reliable NLI dataset.%

Previous work has emphasized the importance of the origin of the train and test data in cross-lingual settings, i.e., whether they are original or created through translation. In particular, \citet{artetxe-etal-2020-translation} show that a mismatch in the origin between training and test data may have a serious impact on the results, particularly when comparing different cross-lingual strategies. Moreover, NLI datasets are known to be biased and contain artifacts that lead models to rely on superficial clues~\citep{gururangan-etal-2018-annotation, poliak-etal-2018-hypothesis, tsuchiya-2018-performance, mccoy-etal-2019-right}. To analyze the impact of these factors in XNLIeu, we have created a Native test set completely from scratch with original premises extracted from sources with content in Basque and hypotheses provided by Basque speakers, which were specifically told to avoid such biases.

Using these datasets, we have conducted a series of experiments using mono- and multilingual language models for Basque, both discriminative and generative, and have tested different training variants for cross-lingual NLI in Basque. The experiments set a new baseline for NLI in Basque, and have served us to analyze the effect of professional post-edition compared to the automatic machine-translation system. We have also identified the most effective cross-lingual strategy for NLI in Basque, considering both translated and native sets.

This paper makes the following contributions:
\begin{itemize}

    \item We develop and release a new dataset for cross-lingual NLI in Basque, which is created by translating the English XNLI, through machine-translation and post-edition. We also release a machine-translated only version of the dataset, as well as a small native dataset for comparison purposes.
    \item We conduct a series of cross-lingual Basque NLI experiments using several language models and following different cross-lingual strategies, and establish new baselines to facilitate research on Basque NLU.
    \item We provide a detailed analysis of the results of our experiments to assess the impact of using different models, strategies and data sources.
\end{itemize}

This paper is structured as follows: Section \ref{sec:related-work} covers some relevant research and resources related to the topic in hand, our dataset is further explained in Section \ref{sec:datasets}, the description of the experiments and experimental settings in Section \ref{sec:experiments}, the results are covered in Section \ref{sec:results}, Section \ref{sec:error-analysis} includes the analysis of the errors in the outputs of our models, Section \ref{sec:conclusion} a summary of the research and its conclusions; and there is a final section that expands on the limitations of our research.

\section{Related work}
\label{sec:related-work}

\paragraph{Cross-lingual NLI.} The best results on NLI benchmarks to date are based on supervised learning, which requires large amounts of training data that are only available for resource-rich languages such as English. Examples of English NLI datasets are the Stanford NLI corpus \citep{snli-bowman-etal-2015-large}, the Multi-genre NLI corpus \citep{mnli-williams-etal-2018-broad} and the Adversarial NLI corpus \citep{nie-etal-2020-adversarial}. The NLI task is also included among the tasks of the popular NLU benchmarks GLUE \citep{wang-etal-2018-glue} and SuperGLUE \citep{wang2020superglue}. Cross-lingual NLI is an alternative approach that leverages pre-trained multilingual models which are fine-tuned in resource-rich languages, then tested in the desired target language. This transfer approach, called \emph{zero-shot}, is often compared to strategies that involve machine translation: \emph{translate-train}, where the training set is translated to each target language and used to train the models on their respective language and \emph{translate-test}, where the test set is translated to the high-resource language, usually English. Alternatively, large multilingual autoregressive models are also known to perform well in cross-lingual settings, by providing them with a set of correct input/label pairs as prompts for new inputs~\citep{brown2020gpt3}.

\paragraph{XNLI.} The Cross-lingual NLI corpus (XNLI) \citep{xnli-conneau-etal-2018} comprises development and test sets in 15 high- and low-resource languages, meant as a cross-lingual benchmark for this task. Later, this corpus was expanded to include additional languages such as Korean \citep{ham-etal-2020-kornli}. 

\paragraph{NLI biases \& artifacts.} Most famous NLI datasets have also been reported to include biases and artifacts \citep{gururangan-etal-2018-annotation,  poliak-etal-2018-hypothesis, tsuchiya-2018-performance, mccoy-etal-2019-right} that should be considered when analyzing the results, as they seem to have critical effects on the performance of systems. \citet{artetxe-etal-2020-translation} analyzes the effect that translated datasets have in cross-lingual settings, due to the so-called \textit{translationese}~\citep{translationese}, and concludes that mismatches between the origin of training and evaluation datasets cause an important impact on the robustness of evaluation.

\paragraph{Evaluation of LLMs.} Nowadays, the focus of the research on evaluation has shifted due to the outstanding growth of LLMs. These models can achieve comparable results to fine-tuned pre-trained models with zero-shot and few-shot approaches for evaluation. Consequently, the focus has shifted towards assessing the models' overall capabilities rather than their performance on specific tasks \citep{llms-survey}. However, low-resource languages like Basque lag behind in NLP development, and can still benefit considerably from semantic datasets for tasks like NLI, which was not previously available for this language.  

\begin{table}[t]
  \centering
  \resizebox{\columnwidth}{!}{%
    \begin{tabular}{lc}
      \hline
      \textbf{Label} & \textbf{Example}                                                                                                          \\ \hline
      premise & \textit{Dena idazten saiatu nintzen}\\
      & \lq{I tried to write everything.}\rq          \\ \hline
      entailment& \makecell{\textit{Nire helburua gauzak idaztea zen.}\\
      \lq{My goal was to write things}\rq}\\
      contradiction& \makecell{\textit{Ez nintzen ezer idazten saiatu ere egin.}\\
      \lq{I didn't even try to write anything.}\rq}\\
      neutral& \makecell{\textit{Aipatu zuen lan bakoitza idatzi nuen.}\\
      \lq{I wrote every paper he mentioned.}\rq}\\ \hline
    \end{tabular}%
  }
  \caption{Examples from the XNLIeu dataset}
  \label{tab:my-table}
\end{table}

\section{The XNLIeu dataset}
\label{sec:datasets}

\textrm{XNLIeu} has been created by machine-translating the English XNLI development and test sets to Basque\footnote{\label{note:elia}All machine translations performed in the paper have been obtained using Elia at \url{https://elia.eus/translator}.} followed by a manual post-edition step\footnote{We hired a professional translation service to perform the post-edition. As is customary, we asked for periodic samples of the post-editions to assert that the translation mistakes from the MT were being corrected and ensure the quality of the post-edited dataset.}. Some examples of \textrm{XNLIeu} are shown in Table \ref{tab:my-table}. We also release the machine-translated version prior to post-edition, dubbed \textrm{XNLIeu$_{\mathrm{MT}}$}, which we use to analyze the effect of post-edition (see Section \ref{sec:results-xnli-xnli}). 

Additionally, we created an original Basque test set from scratch, henceforth referred to as \textit{native}, and compared the results with \textrm{XNLIeu} and \textrm{XNLIeu$_{\mathrm{MT}}$} (see Section \ref{sec:results-native}). Inspired by \citet{snli-bowman-etal-2015-large} and \citet{artetxe-etal-2020-translation}, we performed the following steps to build the native dataset:
\begin{itemize}
    \item As a starting point, we extracted $5,000$ sentences from recent news in Basque, ensuring that they were not previously seen by the models used in the experiments. For this, we scraped Basque News sites and selected sentences from documents whose creation time was posterior to the release date of the pre-training corpora.%
    \item From these initial sentences, we manually selected $207$ sentences that we deemed appropriate for this task, and used them as premises. Examples of phrases that were discarded are headlines, image descriptions that do not include
    verbs, or questions, since it is not always possible to obtain the truth conditions of these types of sentences
    \item We redacted annotation guidelines that explain the task and provide examples to the annotators. In these guidelines, annotators are asked to be creative and to avoid as much as possible some of the annotation artifacts that have been found in the large datasets~\citep{gururangan-etal-2018-annotation, poliak-etal-2018-hypothesis}, such as the use of negation to create contradictions. The detailed guidelines are described in Appendix \ref{sec:guidelines}.
    \item With the assistance of native Basque speakers, one hypothesis was created per premise and label, resulting in three hypotheses per premise, with a total of $621$ sentences.
    \item We performed a final series of minor corrections on the resulting dataset, correcting typos and ensuring that the meaning conveyed by the hypotheses entails the assigned label.
\end{itemize}

\begin{table}[t]
  \centering
\resizebox{\columnwidth}{!}{%
\begin{tabular}{ll|lll}
\hline
                                       & \textbf{XNLI (english)} & \textbf{XNLIeu} & \textbf{XNLIeu$_{\mathrm{MT}}$} & \textbf{native} \\\hline
\textbf{entailment}                    & 9.89          & 8.15            & 7.81                            & 8.95            \\
\textbf{contradiction}                 & 10.39         & 8.73            & 8.39                            & 9.94            \\
\textbf{neutral}                       & 11.4          & 9.31            & 8.98                            & 9.41            \\\hline
\end{tabular}
}
  \caption{Average length of hypotheses for each semantic relation type in our three datasets, as well as the average for the original English instances.}
\label{tab:hyp-len}
\end{table}

Finally, we also distribute a machine-translated version of the English MNLI training corpus to Basque, with a total of $392,702$ sentences, which we use in the translate-train experiments.

\begin{figure*}[t]
\centering
\includegraphics[width=0.3\textwidth]{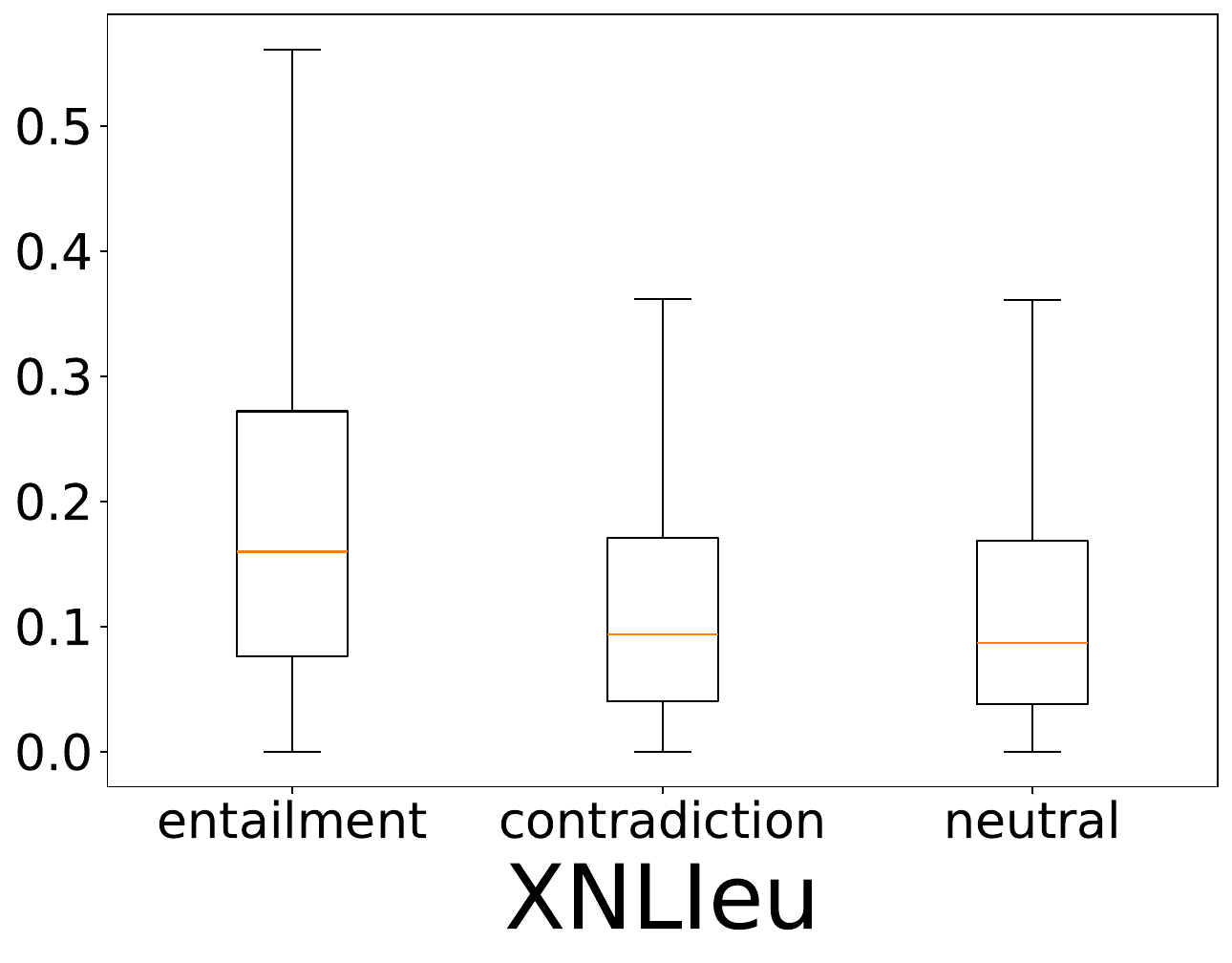}
\includegraphics[width=0.3\textwidth]{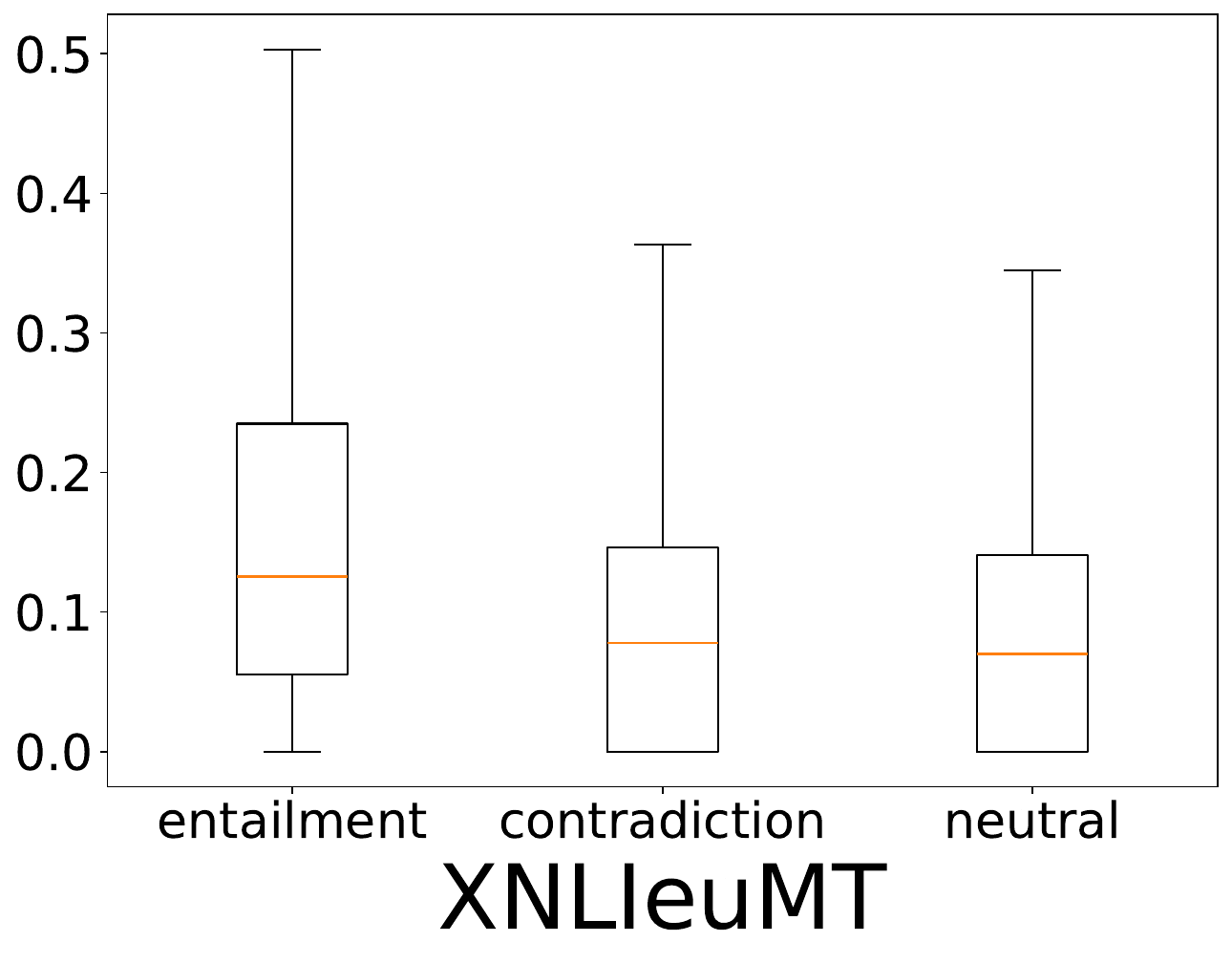}
\includegraphics[width=0.3\textwidth]{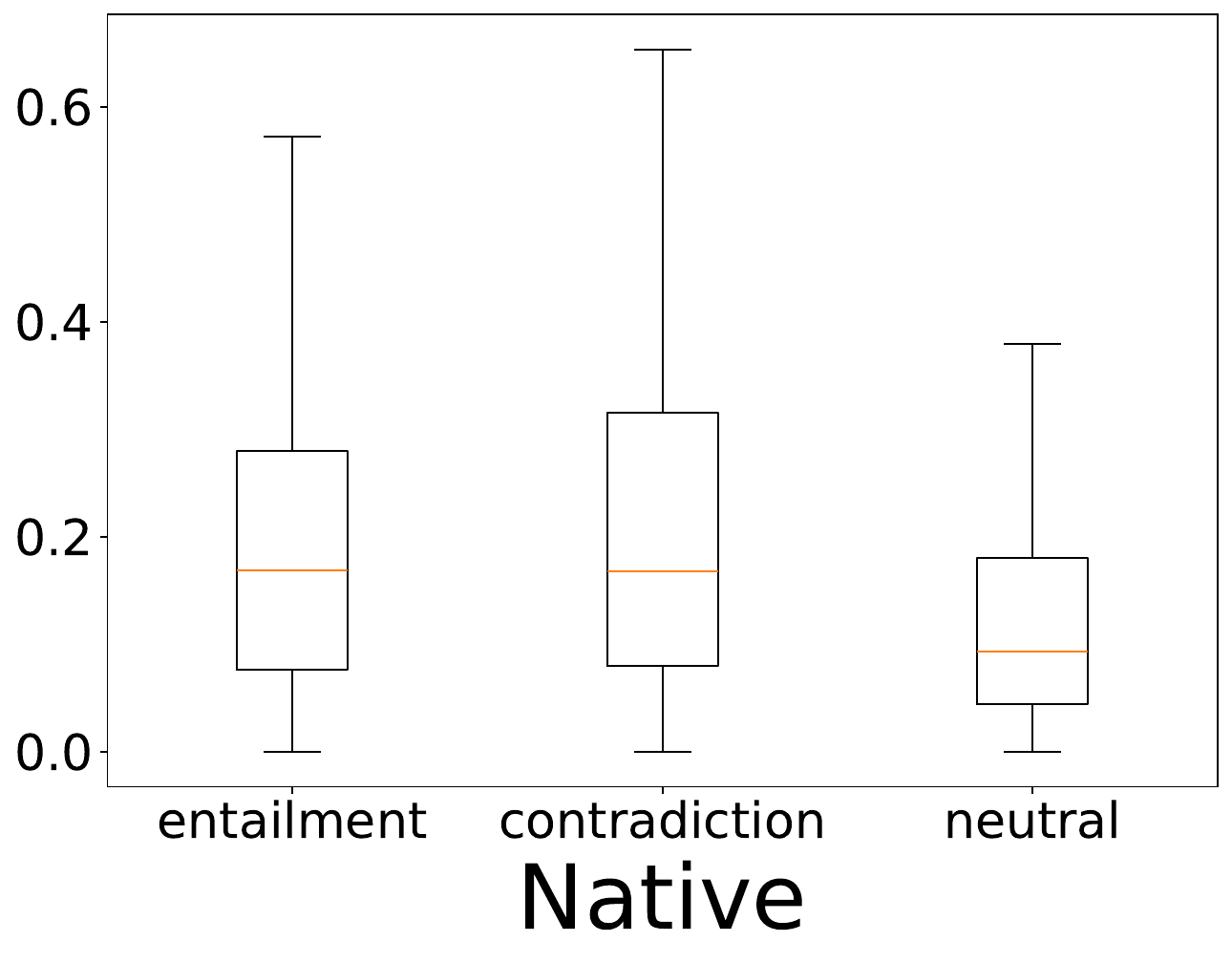}

\caption{Box plots of the lexical overlap between premises and hypotheses calculated with cosine similarity of the three datasets.}
\label{fig:cos-sim}
\centering
\end{figure*}

\subsection{Quantitative analysis}
\label{sec:quant-analys}
In this section we present a quantitative analysis of various aspects of the three developed datasets: \textrm{XNLIeu}, \textrm{XNLIeu$_{\mathrm{MT}}$} and the Native dataset.

\paragraph{Label distribution.} Since there are three hypotheses for each premise in the dataset, the label distribution is perfectly balanced, resulting in no majority class and establishing the baseline accuracy at 33\%. This applies to all three datasets. 

\paragraph{Sentence length.} The average token length for hypotheses for each semantic relation type, as shown in Table \ref{tab:hyp-len}, indicates that there is a bias, as neutral hypotheses are longer on average, while entailed hypotheses tend to be shorter, likely because entailed sentences are often formed by omitting words from the premise \citep{gururangan-etal-2018-annotation}. This bias is present in the original instances in English of the XNLI dataset and in \textrm{XNLIeu} and \textrm{XNLIeu$_{\mathrm{MT}}$}. The hypotheses of the Basque datasets tend to be shorter than the original English ones, but the unbalance between the different semantic relation types is maintained. The native set is also skewed, but in this case, the contradictions are slightly longer than neutral hypotheses, and entailments are still shorter on average.

\paragraph{Word frequency.} Examining word frequency per label is insightful, especially since studies such as \citet{gururangan-etal-2018-annotation} or \citet{tsuchiya-2018-performance} have reported that some NLI datasets exhibit a bias where the contradiction label is strongly associated with negation words. This seems to hold for the \textrm{XNLIeu} and \textrm{XNLIeu$_{\mathrm{MT}}$} datasets. As we can see in Table \ref{tab:frequent-words-english}, the word \textit{ez} \lq{no}\rq~appears much more frequently in contradictions, and so do some other negations like \textit{inork} \lq{nobody}\rq~or \textit{inoiz} \rq{never}\lq. It is plausible that models might be exploiting this feature as a form of shortcut learning for classification without even looking at the premise. The native dataset does not seem to be biased towards negation words, since the guidelines specifically asked the annotators to avoid using artifacts as much as possible (see Appendix \ref{sec:guidelines}). It is interesting to note that among the most frequent words in this dataset, there are frequent references to the Basque culture: \textit{euskaraz} \lq{in Basque}\rq, \textit{euskara} / \textit{euskal} \lq{Basque}\rq or \textit{Bilboko} \lq{from Bilbao}\rq.

\paragraph{Lexical overlap.} The lexical overlap between the premise and hypothesis has been calculated as the cosine similarity between the TF-IDF vector representations of both sentences. The results in Figure \ref{fig:cos-sim} show that in \textrm{XNLIeu} and \textrm{XNLIeu}$_{\mathrm{MT}}$ the highest overlap occurs between premises and entailed hypotheses. This is a known bias in NLI and is attributed to the fact that entailed hypotheses are easy to create by simply omitting parts of the premise~\citep{gururangan-etal-2018-annotation}. In contrast, this bias is not present in the native dataset, where on average the premises overlap mostly with both entailed and contradiction hypotheses, and less with neutral hypotheses.

\begin{table}[t]
\centering
\resizebox{\columnwidth}{!}{%
\begin{tabular}{lllllll}
\hline
                                          & \multicolumn{2}{c}{\textbf{XNLIeu}} & \multicolumn{2}{c}{\textbf{XNLIeu$_{\mathrm{MT}}$}} & \multicolumn{2}{c}{\textbf{native}}                       \\\hline
                                          & no                                  & 0.58\%                                              & no                 & 0.54\% & in Basque          & 0.41\% \\
                                          & \textit{auxiliary}\footnotemark     & 0.24\%                                              & \textit{auxiliary} & 0.23\% & film               & 0.24\% \\
\textbf{entailment}                       & something                           & 0.19\%                                              & some               & 0.18\% & \textit{auxiliary} & 0.24\% \\
                                          & some                                & 0.18\%                                              & something          & 0.16\% & movie              & 0.24\% \\
                                          & \textit{auxiliary}                  & 0.17\%                                              & like               & 0.13\% & of the world       & 0.24\% \\ 
 \hline
                                          & no                                  & 1.61\%                                              & no                 & 1.65\% & no                 & 0.45\% \\
                                          & nobody                              & 0.24\%                                              & nobody             & 0.23\% & in Basque          & 0.34\% \\
\textbf{contradiction}                    & never                               & 0.2\%                                               & \textit{auxiliary} & 0.18\% & Basque             & 0.28\% \\
                                          & \textit{auxiliary}                  & 0.18\%                                              & never              & 0.16\% & my                 & 0.23\% \\
                                          & my                                  & 0.16\%                                              & importance         & 0.14\% & from Bilbao        & 0.23\% \\
 \hline
                                          & no                                  & 0.33\%                                              & no                 & 0.31\% & like               & 0.37\% \\
                                          & my                                  & 0.21\%                                              & dollar             & 0.2\%  & no                 & 0.37\% \\
\textbf{neutral}                          & \textit{auxiliary}                  & 0.19\%                                              & my                 & 0.2\%  & Basque             & 0.25\% \\
                                          & some                                & 0.18\%                                              & \textit{auxiliary} & 0.16\% & sometimes          & 0.25\% \\
                                          & like                                & 0.15\%                                              & some               & 0.16\% & people             & 0.25\% \\ \hline
\end{tabular}
}
\caption{Proportion of most frequent words of the three datasets, translated from Basque to English.}
\label{tab:frequent-words-english}
\end{table}
\footnotetext{Auxiliaries are further discussed in Appendix \ref{sec:app-freq-words}.}

\section{Experimental design}
\label{sec:experiments}

We have conducted a series of experiments on cross-lingual NLI for Basque, using different discriminative and generative language models, both mono- and multilingual. All models have been tested using the three datasets described in Section \ref{sec:datasets}. We aim to determine if post-edition introduces significant changes to the dataset that enhance its reliability. We also want to compare the results on the XNLI-derived datasets with the native human-devised dataset, and analyze the effect of biases and artifacts introduced by translation. Since there is no training set in Basque for NLI, we consider different cross-lingual alternatives\footnote{The translate-test approach has not been implemented since the datasets have been originally translated from English to Basque, so back-translating them to English would not allow us to draw meaningful conclusions.}: 
\begin{itemize}
    \item \emph{Zero-Shot transfer}: We use multilingual discriminative models that have been pre-trained at least in English and Basque. These models are then fine-tuned on the English MNLI corpus. In a further experiment, we explore fine-tuning with source languages beyond English.
    \item \emph{Translate-train}: We machine-translate the English MNLI dataset to Basque, and use it to fine-tune the discriminative models (both multilingual and Basque monolingual).
    \item \emph{Zero-shot prompting}: We directly test multilingual generative models that include Basque, without fine-tuning. We prompt the models by combining the premise and the hypothesis according to a template that is different for each possible label (See Appendix \ref{sec:prompts}). %
\end{itemize}

\begin{table}[t]
\centering
\resizebox{\columnwidth}{!}{%
\begin{tabular}{lcc}
\hline
 \multicolumn{3}{c}{\textbf{Discriminative}}                                                                                \\
\hline
  \multicolumn{1}{c}{\textbf{Name}} & \multicolumn{1}{c}{\textbf{Language}} & \multicolumn{1}{c}{\textbf{\# of parameters}} \\ \hline
  IXAmBERT                          & Multilingual                                 & 177M                                          \\
  multilingual BERT                 & Multilingual                                 & 179M                                          \\
  XLM-RoBERTa (base)                & Multilingual                                 & 279M                                          \\
  XLM-RoBERTa (large)               & Multilingual                                 & 561M                                          \\
  BERTeus                           & Basque                                & 124M                                          \\
  RoBERTa-eus Euscrawl              & Basque                                & 355M                                          \\\hline
  \multicolumn{3}{c}{\textbf{Generative}}                                                                                   \\\hline
  Latxa                             & Multilingual                                 & 7B                                            \\ 
  BLOOM                             & Multilingual                                 & 7.1B                                          \\
  XGLM                              & Multilingual                                 & 7.5B \\\hline
\end{tabular}
}
\caption{Details of the models used in the experiments.}
\label{tab:models}
\end{table}

Regarding the models, we have experimented with the following discriminative models: IXAmBERT \citep{ixambert}, multilingual BERT \citep{bert}, XLM-RoBERTa large \citep{roberta}, BERTeus \citep{berteus} and RoBERTa-eus-large \citep{euscrawl-artetxe2022does}. Further details about these models can be found in Table \ref{tab:models}. All of the models have been used in their cased version. For the BERT models, we have used a learning rate of 5e\textsuperscript{-5}, and for the RoBERTa models, we have used a smaller learning rate of 10e\textsuperscript{-6}, which is the only hyperparameter that has not been kept default, to avoid a degenerated solution. All models have been trained for 10 epochs, and the model selection has been performed on the development test. There has been no further attempt at hyperparameter optimization, since the goal was not to obtain the best possible model, but rather to compare the effects of the different sets and strategies. The models have been trained with three different random seeds to get the mean and the standard deviation and reduce the effects of randomness associated with initializing the weights and selecting the order of the training data. The code used for the experiments with discriminative models has been adapted from the code examples for fine-tuning for different tasks provided by \citet{wolf-etal-2020-transformers}. 

We have also tested three multilingual generative models that include Basque among their pre-training languages: BLOOM \citep{workshop2023bloom}, XGLM \citep{xglm-lin2022fewshot} and Latxa \citep{etxaniz2024latxa}, a model based on Llama 2 tuned for Basque with continual pretraining on Basque corpora. The prompts used in our experiments can be seen in Appendix \ref{sec:prompts}. As for evaluation, we select the label whose log-likelihood is highest, according to the model. The code used for testing the generative models is based on that included in the Language Model Evaluation Harness project \citep{eval-harness}.

Following usual practice, we use accuracy as our evaluation metric: the ratio of correctly classified instances divided by the total number of instances.

\section{Results}
\label{sec:results} 

\begin{table}[t]
  \centering
\resizebox{\columnwidth}{!}{%
\begin{tabular}{lll}
\hline
\multicolumn{3}{c}{\textbf{zero-shot}}                                                                                               \\ \hline
 \textbf{}                              & \textbf{XNLIeu}                 & \textbf{XNLIeu$_{\mathrm{MT}}$}                           \\ \hline
 \textbf{IXAmBERT}                      & 72.5 ($\pm1.4e^{-3}$)           & 67.3 ($\pm7.0e^{-3}$)\\
 \textbf{mBERT}                         & 60.1 ($\pm5.7e^{-3}$)           & 57.9 ($\pm1.2e^{-2}$)                  \\
 \textbf{XLM-RoBERTa base}              & 73.4 ($\pm3.5e^{-3}$)           & 69.0 ($\pm9.0e^{-3}$)\\
 \textbf{XLM-RoBERTa large}             & \textbf{81.1} ($\pm2.8e^{-3}$)& \textbf{75.4} ($\pm2.0e^{-3}$)\\\hline

\multicolumn{3}{c}{\textbf{translate-train}} \\ \hline
\textbf{}                                & \textbf{XNLIeu}             & \textbf{XNLIeu$_{\mathrm{MT}}$}                          \\ \hline
  \textbf{IXAmBERT}                       & 75.9 ($\pm6.4e^{-3}$)        & 71.3 ($\pm4e^{-3}$)\\
  \textbf{mBERT}                          & 74.8 ($\pm4.2e^{-3}$)        & 71.3 ($\pm0.0$)\\
  \textbf{XLM-RoBERTa large}              & \textbf{83.8} ($\pm6.0e^{-4}$)& \textbf{79.9} ($\pm1.0e^{-3}$)\\
  \textbf{RoBERTa-euscrawl}                & 83.0 ($\pm7.1e^{-3}$)& 78.6 ($\pm2.0e^{-3}$)\\ 
  \textbf{BERTeus}                         & 79.0 ($\pm4.2e^{-3}$)& 74.9 ($\pm8.0e^{-3}$)\\
\hline
\end{tabular}
}
\caption{Accuracy of discriminative fine-tuned models tested with \textrm{XNLIeu} and XNLIeu$_{\mathrm{MT}}$ datasets (mean and standard deviation of three runs). Best results in bold.}
\label{tab:discriminative-result}
\end{table}

In this section, we show the main results of our experiments and discuss the main findings. We start by analyzing the results on the datasets derived from XNLI (XNLIeu and XNLIeu$_{\mathrm{MT}}$), followed by a comparison with those obtained using the native dataset. Finally, we detail the results of experiments that involved fine-tuning with source languages other than English.

\subsection{Results for XNLIeu and XNLIeu$_{\mathrm{MT}}$}
\label{sec:results-xnli-xnli}

The main results for the discriminative models can be seen in Table \ref{tab:discriminative-result}. 
All systems perform consistently better when evaluated on the post-edited \textrm{XNLIeu} compared to the machine-translated XNLIeu$_{\mathrm{MT}}$, and in some cases, the relative ranking among the models change, as is the case between multilingual BERT and IXAmBERT in the translate-test setting. %
Translate-train obtains better results overall on all models, and the difference is slightly higher in the XNLIeu$_{\mathrm{MT}}$ dataset ($7.3\%$ accuracy points on average), where both training and test data have been created only through machine-translation. This result is consistent with the findings reported in \citet{artetxe-etal-2020-translation}. Multilingual BERT is the model that improves the most with translate-train, probably because the presence of Basque at pre-training time was lower compared to the other models.

\begin{table}[t]
\centering
\begin{tabular}{lll}
\hline
\textbf{}                                & \textbf{XNLIeu} & \textbf{XNLIeu$_{\mathrm{MT}}$}         \\ \hline
\textbf{Latxa}                       & \textbf{50.9}            & \textbf{47.8}\\
\textbf{BLOOM}                           & 49.5& 47.5\\
\textbf{XGLM}                            & 48.1                       & 46.7                                  \\ \hline

\end{tabular}
\caption{Accuracy of generative models tested with \textrm{XNLIeu} and XNLIeu$_{\mathrm{MT}}$ datasets using a zero-shot prompting approach. Best results in bold.}
\label{tab:generative-results}
\end{table}

Table \ref{tab:generative-results} shows the results obtained by the generative models. Once again, the models perform better when evaluated on the post-edited XNLIeu, but the performance gap is smaller compared with fine-tuned approaches. In any case, the results suggest that post-edition introduces significant changes to the dataset and is therefore important in order to obtain a reliable evaluation benchmark. We analyze this aspect further in Section \ref{sec:error-analysis}.

\subsection{Results for the native test set}
\label{sec:results-native}

Table \ref{tab:native-results} shows the results of the models when evaluated on the native dataset. %
The translate-train approach still yields better results than zero-shot transfer, but the difference in accuracy between both approaches is on average $2\%$ percentage points smaller than those obtained with the translated sets. This is likely a consequence of the mismatch between the train and test sets, because in this setting, the former is built through translation text while the latter is natively written in Basque.

Discriminative models perform worse on the native dataset, with approximately 10\% lower accuracy on average. While comparing results among different datasets is not always meaningful, we attribute the performance drop to the fact that the native dataset is less biased, as seen in Section ~\ref{sec:quant-analys}. As a consequence, the models cannot rely on superficial patterns to deduce the relation between sentences, which makes this dataset especially challenging. Another possible cause is the notable presence of references to the Basque culture as it was sourced from original Basque materials.

\begin{table}[t]
    \centering
    \begin{tabular}{ll}
      \hline
      \multicolumn{2}{c}{\textbf{zero-shot transfer}}\\\hline
      \textbf{IXAmBERT}          &64.0 ($\pm9.0e^{-3}$)\\
      \textbf{mBERT}             &52.4 ($\pm1.6e^{-2}$)\\
      \textbf{XLM-RoBERTa base}  &65.3 ($\pm7.0e^{-3}$)\\
      \textbf{XLM-RoBERTa large} &\textbf{73.8} ($\pm7.0e^{-3}$)\\\hline
      \multicolumn{2}{c}{\textbf{translate-train}}\\\hline
      \textbf{BERTeus}           &68.4 ($\pm1.0e^{-2}$)\\
      \textbf{IXAmBERT}          &65.6 ($\pm1.0e^{-2}$)\\
      \textbf{mBERT}             &62.8 ($\pm9.0e^{-3}$)\\
      \textbf{RoBERTa-euscrawl}  &75.2 ($\pm7.0e^{-3}$)\\
      \textbf{XLM-RoBERTa large} &\textbf{76.4} ($\pm1.3e^{-2}$)\\\hline
      \multicolumn{2}{c}{\textbf{zero-shot prompting}}\\\hline
      \textbf{Latxa}&\textbf{53.3}\\
      \textbf{BLOOM}&49.8\\
      \textbf{XGLM}&46.5\\\hline
    \end{tabular}
    \caption{Accuracy of discriminative (upper part) and generative (bottom part) models tested on the native dataset. Best results in bold.}
    \label{tab:native-results}
\end{table}

Generative models yield results that are comparable to those obtained with machine-translated and post-edited sets. This result is a consequence of the zero-shot prompting strategy followed in generative models, which does not include fine-tuning, and therefore does not rely on examples that can induce bias in the model.

\subsection{Choice of the source language}
\label{sec:choice-source-lang}

\begin{table}[t]
\centering
\resizebox{\columnwidth}{!}{%
\begin{tabular}{clll}
\hline
\multicolumn{1}{l}{} & \multicolumn{1}{c}{\textbf{XNLIeu}} & \multicolumn{1}{c}{\textbf{XNLIeu$_{\mathrm{MT}}$}} & \multicolumn{1}{c}{\textbf{native}} \\\hline
\textbf{en}          & 73.4 ($\pm3.5e^{-3}$)               & 69.0 ($\pm9.0e^{-3}$)                               & \textbf{65.3 ($\pm7.0e^{-3}$)}      \\ \hline
\textbf{ar}          & \textbf{73.9} ($\pm2.6e^{-3}$)       & \textbf{71.2} ($\pm4.0e^{-3}$)                       & 61.9 ($\pm3.0e^{-3}$)                \\
\textbf{bg}          & 73.2 ($\pm8.9e^{-3}$)                & \underline{71.0} ($\pm2.1e^{-3}$)                    & 62.7 ($\pm9.0e^{-3}$)                \\
\textbf{de}          & \underline{73.9} ($\pm5.3e^{-3}$)    & 70.4 ($\pm7.0e^{-4}$)                                & 63.5 ($\pm8.0e^{-3}$)                \\
\textbf{el}          & 73.7 ($\pm1.7e^{-3}$)                & 70.7 ($\pm7.0e^{-4}$)                                & 63.6 ($\pm7.0e^{-3}$)                \\
\textbf{es}          & 73.7 ($\pm5.2e^{-3}$)                & 70.3 ($\pm7.0e^{-4}$)                                & \underline{65.0} ($\pm7.0e^{-3}$)    \\
\textbf{fr}          & 73.7 ($\pm4.9e^{-3}$)                & 69.9 ($\pm7.1e^{-3}$)                                & 63.3 ($\pm2.1e^{-2}$)                \\
\textbf{hi}          & 73.3 ($\pm7.0e^{-3}$)                & 70.7 ($\pm4.2e^{-3}$)                                & 62.3 ($\pm5.0e^{-3}$)                \\
\textbf{ru}          & 72.9 ($\pm1.5e^{-3}$)                & 69.7 ($\pm2.1e^{-3}$)                                & 62.2 ($\pm6.0e^{-3}$)                \\
\textbf{sw}          & 71.8 ($\pm3.1e^{-3}$)                & 68.3 ($\pm7.1e^{-3}$)                                & 63.1 ($\pm6.0e^{-3}$)                \\
\textbf{th}          & 73.0 ($\pm6.7e^{-3}$)                & 70.2 ($\pm4.2e^{-3}$)                                & 64.1 ($\pm6.0e^{-3}$)                \\
\textbf{tr}          & 73.5 ($\pm6.2e^{-3}$)                & 70.9 ($\pm7.0e^{-4}$)                                & 63.6 ($\pm7.0e^{-3}$)                \\
\textbf{ur}          & 66.5 ($\pm4.6e^{-3}$)                & 65.0 ($\pm1.4e^{-3}$)                                & 56.0 ($\pm1.1e^{-2}$)                \\
\textbf{vi}          & 72.6 ($\pm1.1e^{-2}$)                & 69.6 ($\pm7.8e^{-3}$)                                & 62.4 ($\pm1.5e^{-2}$)                \\
\textbf{zh}          & 71.8 ($\pm7.0e^{-3}$)                & 69.7 ($\pm2.1e^{-3}$)                                & 62.0 ($\pm6.0e^{-3}$)                \\ \hline
\end{tabular}
}
\caption{Zero-shot cross-lingual transfer accuracy of XLMRoBERTa fine-tuned in different languages (mean and standard deviation of three runs). Best results in bold, second best underlined.}
\label{tab:zero-shot-languages}
\end{table}
We have conducted additional typological experiments to test the impact of the choice of the source language in a zero-shot cross-lingual transfer setting for Basque. For this, we fine-tuned XLM-RoBERTa-base in $14$ languages using machine-translated versions of the MNLI training data, as well as English, and tested them on XNLIeu, XNLIeu$_{\mathrm{MT}}$ and the native test set. The results of these experiments are depicted in Table \ref{tab:zero-shot-languages}. 

The table shows small differences in XNLIeu and XNLIeu$_{\mathrm{MT}}$. We attribute these results to the fact that in this setting, both the training and test data come from translations, which lessens the importance of which source language to use. This is not the case for English, whose train data is original and not translated, but still it is not among the languages that achieve the highest results. When tested on the native dataset factors such as proximity between languages and loanword frequency gain relevance, as shown in the table, and the difference among languages is higher. Choosing English or Spanish yields similar results, while the performance when any other language is selected is noticeably lower.

\begin{figure*}[t!]

         \includegraphics[width=0.3\textwidth]{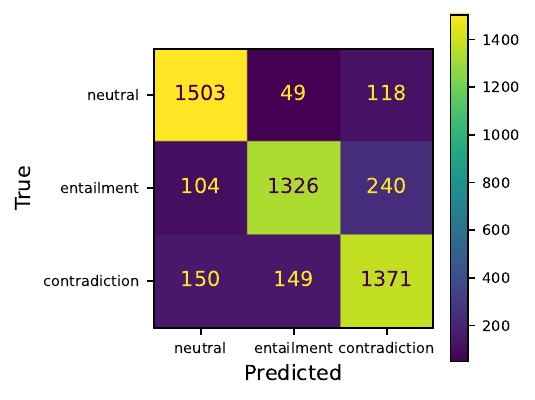}
         \includegraphics[width=0.3\textwidth]{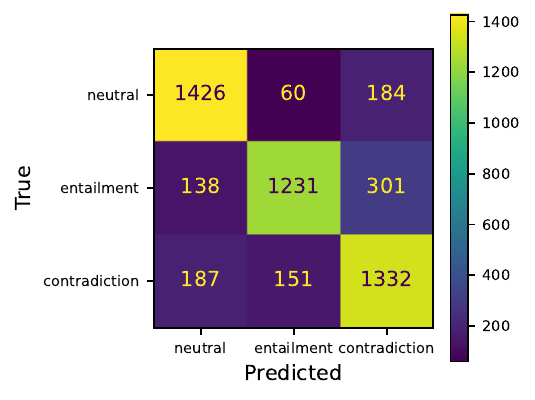}
         \includegraphics[width=0.3\textwidth]{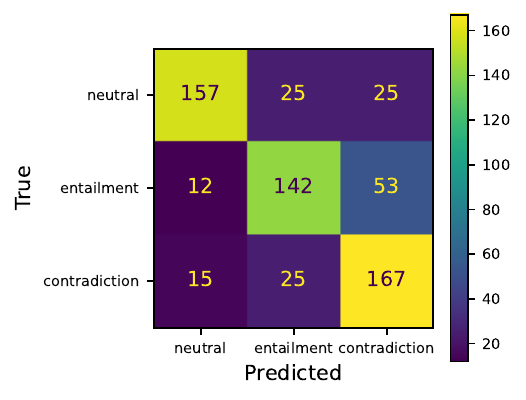}
         
            \begin{small}
                \hspace{21mm}XNLIeu
                \hspace{38mm}XNLIeu$_{\mathrm{MT}}$
                \hspace{35mm}Native
            \end{small}
        \caption{Confusion matrices for the XLM-RoBERTa large fine-tuned in Basque, our best model, tested in our three datasets. Best viewed in color.}

        \label{fig:conf-matr}
\end{figure*}

\section{Analysis}
\label{sec:error-analysis}

This Section provides additional analyses of the results. We begin by considering the performance of the best model on a per-label basis, followed by a manual comparison of the model outputs on the XNLIeu and XNLIeu$_{\mathrm{TM}}$ datasets to analyze the effects of post-edition.

\subsection{Results per label}

Figure \ref{fig:conf-matr} shows the confusion matrices on each label (entailment/neutral/contradiction) corresponding to the model and setting that performed best, XLM-RoBERTa large fine-tuned in Basque. For both XNLIeu and XNLIeu$_{\mathrm{MT}}$, the label that gets the higher F1 score is contradiction ($87.7$ and $83.4$ respectively), followed by entailment ($83$ and $79.1$), while neutral instances obtain the worst F1 score overall ($80.7$ and $76.4$). This is in accordance with the analysis performed in Section~\ref{sec:quant-analys}, which indicates the presence of biases in these datasets, as well as in the training dataset. The results suggest that the models do rely on those biases, for instance by classifying instances where the hypothesis contains negative words as contradictions, or those where the hypothesis is short and has large lexical similarity with premises as entailment. On the other hand, no specific biases were detected in neutral instances, and consequently, it is more difficult for models to correctly classify them.

Section~\ref{sec:quant-analys} reveals that the native dataset does not suffer from such apparent biases, and this is again reflected in the results depicted in Figure \ref{fig:conf-matr} for this dataset (right part). While contradiction is still the label with the best F1 score ($80.3$), now the label that attains the worst F1 is entailment ($71.2$), and the second-best is neutral ($73.9$).

\subsection{Effects of post-edition }

Section \ref{sec:results} reveals that systems perform consistently worse when evaluated on the machine-translated \textrm{XNLIeu$_{\mathrm{MT}}$} dataset compared to the post-edited \textrm{XNLIeu}. So as to get a deeper insight into this result, we performed an analysis on \textrm{XNLIeu} and \textrm{XNLIeu$_{\mathrm{MT}}$} by selecting instances that have been correctly predicted in one dataset and wrongly predicted in the other. The analysis reveals that \textrm{XNLIeu$_{\mathrm{MT}}$} often contains translation errors that change the relation between premise and hypothesis, and that when post-editing the professional translators corrected those errors. The most frequent error converts entailment and contradiction hypotheses to neutral. Common translation errors include:

\begin{itemize}
 \item Changing the polarity of a sentence from negative to positive or vice versa.
\ex
 Original: No, I live off campus.\\
 MT: \textit{ez naiz campusetik kanpo bizi}\\
 \lq{I don't live off campus}\rq
 \xe
 \item Using an incorrect auxiliary verb, which can have a detrimental effect and completely change the meaning of a sentence.
  \ex
 Original: I was still scared.\\
 MT: \textit{eta oraindik beldurra ematen dit}\\
 \lq{I am still scared}\rq
\xe
\item Omitting crucial information from the original sentence or occasionally creating nonsensical sentences. 
 \ex
 Original: I like feeling myself.\\
 MT: \textit{Nik neuk gustuko dut ontzia.}\\
 \lq{I like the vessel myself}\rq
\xe
\end{itemize}

On the other hand, there do not seem to be clear patterns in those instances that have been correctly predicted on \textrm{XNLIeu$_{\mathrm{MT}}$} and incorrectly on \textrm{XNLIeu}. We have only found a handful of examples where the original label of XNLI is ambiguous and post-edition introduces necessary changes to make the translations accurate and fluent, which can alter the relation between both sentences.

\section{Conclusions}
\label{sec:conclusion}

In this work, we introduce XNLIeu, a new dataset for cross-lingual NLI in Basque. XNLIeu is developed by machine-translating the English part of XNLI followed by a post-edition step with the assistance of professional translators. Along with XNLIeu we release the full machine-translated version, as well as a Basque native version carefully built to avoid known biases in NLI datasets. We have conducted a series of cross-lingual Basque NLI experiments using a set of language models and different cross-lingual strategies. The experiments show that translate-train is the best strategy, particularly when there is no mismatch between the origin of the train and test data. In the native dataset, translate-train still yields the best results, but the difference is comparatively smaller. This finding aligns with prior research examining the effects of translation-based datasets. We also manually analyze the results of the models and find that machine-translation often introduces artifacts that change the meaning of the premises or hypotheses, and that professional translators correct those errors when post-editing. We conclude that post-edition is a crucial step towards reliable evaluation of cross-lingual NLI.

All of the datasets developed in this paper are publicly available under the same licenses as XNLI. We believe that they are an important resource that will contribute to filling the gaps in resources that exist in Basque, which can hinder the development of research and applications with a focus on semantics in this language.

\section*{Limitations}
\label{sec:limitations}
Some limitations to this study should be taken into account, specially in the design of future research. 

We have centered our work around the Basque language, which is considered to be a low-resource language. This means that, although some LLMs feature Basque in their training, there is not as much data and tools available as for other languages like English or Spanish. This was the main motivation for this research, but there is no prior work about NLI in Basque to be used as a reference, specifically in the experimental design and the interpretation of the results of the experiments. 

Generative models are becoming more complex and versatile and are currently a popular subject of investigation. Most modern evaluation approaches are not focused on creating large corpora for specific tasks, but rather on testing generative models using prompt engineering and zero-shot or few-shot strategies. Our approach may seem outdated, as our research has focused mainly on the creation of our datasets and discriminative models, and generative models have only been tested with a zero-shot prompting approach.  Future research for NLI in Basque should extend this line of research to account for the most recent developments and should include more insight into effective prompts and experiments performed with strategies other than zero-shot. However, we believe that the creation of our dataset and the approach we have followed are still pertinent for a low-resource language like Basque, which unfortunately does not include all the necessary resources to fully leverage the most recent advances brought by generative models, and can take advantage of a task like NLI, which enables the development of semantic applications and is useful for transfer-learning into a lot of different tasks. 

\section*{Acknowledgements}

This work has been partially supported by the Basque Government (Research group funding IT-1805-22 and ICL4LANG project, grant no. KK-2023/00094) as well as the DeepR3 project (TED2021-130295B-C31) founded by MCIN/AEI/10.13039/501100011033 and European Union NextGeneration EU/PRTR. Julen Etxaniz holds a PhD grant from the Basque Government (PRE\_2023\_2\_0060).

\bibliography{bibliography}

\begin{thebibliography}{29}
\expandafter\ifx\csname natexlab\endcsname\relax\def\natexlab#1{#1}\fi

\bibitem[{Agerri et~al.(2020)Agerri, Vicente, Campos, Barrena, Saralegi, Soroa, and Agirre}]{berteus}
Rodrigo Agerri, Iñaki~San Vicente, Jon~Ander Campos, Ander Barrena, Xabier Saralegi, Aitor Soroa, and Eneko Agirre. 2020.
\newblock \href {http://arxiv.org/abs/2004.00033} {Give your text representation models some love: the case for basque}.

\bibitem[{Artetxe et~al.(2022)Artetxe, Aldabe, Agerri, Perez-de Vi{\~n}aspre, and Soroa}]{euscrawl-artetxe2022does}
Mikel Artetxe, Itziar Aldabe, Rodrigo Agerri, Olatz Perez-de Vi{\~n}aspre, and Aitor Soroa. 2022.
\newblock \href {https://doi.org/10.18653/v1/2022.emnlp-main.499} {Does corpus quality really matter for low-resource languages?}
\newblock In \emph{Proceedings of the 2022 Conference on Empirical Methods in Natural Language Processing}, pages 7383--7390, Abu Dhabi, United Arab Emirates. Association for Computational Linguistics.

\bibitem[{Artetxe et~al.(2020)Artetxe, Labaka, and Agirre}]{artetxe-etal-2020-translation}
Mikel Artetxe, Gorka Labaka, and Eneko Agirre. 2020.
\newblock \href {https://doi.org/10.18653/v1/2020.emnlp-main.618} {Translation artifacts in cross-lingual transfer learning}.
\newblock In \emph{Proceedings of the 2020 Conference on Empirical Methods in Natural Language Processing (EMNLP)}, pages 7674--7684, Online. Association for Computational Linguistics.

\bibitem[{{BigScience Workshop} et~al.(2023){BigScience Workshop}, Scao, Fan, Akiki, Pavlick, Ili{\'c}, Hesslow, Castagn{\'e}, Luccioni, Yvon et~al.}]{workshop2023bloom}
{BigScience Workshop}, Teven~Le Scao, Angela Fan, Christopher Akiki, Ellie Pavlick, Suzana Ili{\'c}, Daniel Hesslow, Roman Castagn{\'e}, Alexandra~Sasha Luccioni, Fran{\c{c}}ois Yvon, et~al. 2023.
\newblock \href {http://arxiv.org/abs/2211.05100} {Bloom: A 176b-parameter open-access multilingual language model}.

\bibitem[{Bowman et~al.(2015)Bowman, Angeli, Potts, and Manning}]{snli-bowman-etal-2015-large}
Samuel~R. Bowman, Gabor Angeli, Christopher Potts, and Christopher~D. Manning. 2015.
\newblock \href {https://doi.org/10.18653/v1/D15-1075} {A large annotated corpus for learning natural language inference}.
\newblock In \emph{Proceedings of the 2015 Conference on Empirical Methods in Natural Language Processing}, pages 632--642, Lisbon, Portugal. Association for Computational Linguistics.

\bibitem[{Brown et~al.(2020)Brown, Mann, Ryder, Subbiah, Kaplan, Dhariwal, Neelakantan, Shyam, Sastry, Askell, Agarwal, Herbert-Voss, Krueger, Henighan, Child, Ramesh, Ziegler, Wu, Winter, Hesse, Chen, Sigler, Litwin, Gray, Chess, Clark, Berner, McCandlish, Radford, Sutskever, and Amodei}]{brown2020gpt3}
Tom Brown, Benjamin Mann, Nick Ryder, Melanie Subbiah, Jared~D Kaplan, Prafulla Dhariwal, Arvind Neelakantan, Pranav Shyam, Girish Sastry, Amanda Askell, Sandhini Agarwal, Ariel Herbert-Voss, Gretchen Krueger, Tom Henighan, Rewon Child, Aditya Ramesh, Daniel Ziegler, Jeffrey Wu, Clemens Winter, Chris Hesse, Mark Chen, Eric Sigler, Mateusz Litwin, Scott Gray, Benjamin Chess, Jack Clark, Christopher Berner, Sam McCandlish, Alec Radford, Ilya Sutskever, and Dario Amodei. 2020.
\newblock Language models are few-shot learners.
\newblock In \emph{Advances in Neural Information Processing Systems}, volume~33, pages 1877--1901. Curran Associates, Inc.

\bibitem[{Conneau et~al.(2019)Conneau, Khandelwal, Goyal, Chaudhary, Wenzek, Guzm{\'{a}}n, Grave, Ott, Zettlemoyer, and Stoyanov}]{roberta}
Alexis Conneau, Kartikay Khandelwal, Naman Goyal, Vishrav Chaudhary, Guillaume Wenzek, Francisco Guzm{\'{a}}n, Edouard Grave, Myle Ott, Luke Zettlemoyer, and Veselin Stoyanov. 2019.
\newblock \href {http://arxiv.org/abs/1911.02116} {Unsupervised cross-lingual representation learning at scale}.
\newblock \emph{CoRR}, abs/1911.02116.

\bibitem[{Conneau et~al.(2018)Conneau, Rinott, Lample, Williams, Bowman, Schwenk, and Stoyanov}]{xnli-conneau-etal-2018}
Alexis Conneau, Ruty Rinott, Guillaume Lample, Adina Williams, Samuel Bowman, Holger Schwenk, and Veselin Stoyanov. 2018.
\newblock \href {https://doi.org/10.18653/v1/D18-1269} {{XNLI}: Evaluating cross-lingual sentence representations}.
\newblock In \emph{Proceedings of the 2018 Conference on Empirical Methods in Natural Language Processing}, pages 2475--2485, Brussels, Belgium. Association for Computational Linguistics.

\bibitem[{Devlin et~al.(2018)Devlin, Chang, Lee, and Toutanova}]{bert}
Jacob Devlin, Ming{-}Wei Chang, Kenton Lee, and Kristina Toutanova. 2018.
\newblock \href {http://arxiv.org/abs/1810.04805} {{BERT:} pre-training of deep bidirectional transformers for language understanding}.
\newblock \emph{CoRR}, abs/1810.04805.

\bibitem[{Dušek et~al.(2023)Dušek, Wawer, Galias, and Wojciechowska}]{retrieval-dusek2023improving}
Roman Dušek, Aleksander Wawer, Christopher Galias, and Lidia Wojciechowska. 2023.
\newblock \href {http://arxiv.org/abs/2308.03103} {Improving domain-specific retrieval by nli fine-tuning}.

\bibitem[{Etxaniz et~al.(2024)Etxaniz, Sainz, Miguel, Aldabe, Rigau, Agirre, Ormazabal, Artetxe, and Soroa}]{etxaniz2024latxa}
Julen Etxaniz, Oscar Sainz, Naiara~Perez Miguel, Itziar Aldabe, German Rigau, Eneko Agirre, Aitor Ormazabal, Mikel Artetxe, and Aitor Soroa. 2024.
\newblock Latxa: An open language model and evaluation suite for basque.
\newblock \emph{arXiv preprint}.

\bibitem[{Gao et~al.(2021)Gao, Tow, Biderman, Black, DiPofi, Foster, Golding, Hsu, McDonell, Muennighoff, Phang, Reynolds, Tang, Thite, Wang, Wang, and Zou}]{eval-harness}
Leo Gao, Jonathan Tow, Stella Biderman, Sid Black, Anthony DiPofi, Charles Foster, Laurence Golding, Jeffrey Hsu, Kyle McDonell, Niklas Muennighoff, Jason Phang, Laria Reynolds, Eric Tang, Anish Thite, Ben Wang, Kevin Wang, and Andy Zou. 2021.
\newblock \href {https://doi.org/10.5281/zenodo.5371628} {A framework for few-shot language model evaluation}.

\bibitem[{Guo et~al.(2023)Guo, Jin, Liu, Huang, Shi, Supryadi, Yu, Liu, Li, Xiong, and Xiong}]{llms-survey}
Zishan Guo, Renren Jin, Chuang Liu, Yufei Huang, Dan Shi, Supryadi, Linhao Yu, Yan Liu, Jiaxuan Li, Bojian Xiong, and Deyi Xiong. 2023.
\newblock \href {http://arxiv.org/abs/2310.19736} {Evaluating large language models: A comprehensive survey}.

\bibitem[{Gururangan et~al.(2018)Gururangan, Swayamdipta, Levy, Schwartz, Bowman, and Smith}]{gururangan-etal-2018-annotation}
Suchin Gururangan, Swabha Swayamdipta, Omer Levy, Roy Schwartz, Samuel Bowman, and Noah~A. Smith. 2018.
\newblock \href {https://doi.org/10.18653/v1/N18-2017} {Annotation artifacts in natural language inference data}.
\newblock In \emph{Proceedings of the 2018 Conference of the North {A}merican Chapter of the Association for Computational Linguistics: Human Language Technologies, Volume 2 (Short Papers)}, pages 107--112, New Orleans, Louisiana. Association for Computational Linguistics.

\bibitem[{Ham et~al.(2020)Ham, Choe, Park, Choi, and Soh}]{ham-etal-2020-kornli}
Jiyeon Ham, Yo~Joong Choe, Kyubyong Park, Ilji Choi, and Hyungjoon Soh. 2020.
\newblock \href {https://doi.org/10.18653/v1/2020.findings-emnlp.39} {{K}or{NLI} and {K}or{STS}: New benchmark datasets for {K}orean natural language understanding}.
\newblock In \emph{Findings of the Association for Computational Linguistics: EMNLP 2020}, pages 422--430, Online. Association for Computational Linguistics.

\bibitem[{Ji et~al.(2023)Ji, Lee, Frieske, Yu, Su, Xu, Ishii, Bang, Madotto, and Fung}]{10.1145/3571730}
Ziwei Ji, Nayeon Lee, Rita Frieske, Tiezheng Yu, Dan Su, Yan Xu, Etsuko Ishii, Ye~Jin Bang, Andrea Madotto, and Pascale Fung. 2023.
\newblock \href {https://doi.org/10.1145/3571730} {Survey of hallucination in natural language generation}.
\newblock \emph{ACM Comput. Surv.}, 55(12).

\bibitem[{Lin et~al.(2022)Lin, Mihaylov, Artetxe, Wang, Chen, Simig, Ott, Goyal, Bhosale, Du, Pasunuru, Shleifer, Koura, Chaudhary, O'Horo, Wang, Zettlemoyer, Kozareva, Diab, Stoyanov, and Li}]{xglm-lin2022fewshot}
Xi~Victoria Lin, Todor Mihaylov, Mikel Artetxe, Tianlu Wang, Shuohui Chen, Daniel Simig, Myle Ott, Naman Goyal, Shruti Bhosale, Jingfei Du, Ramakanth Pasunuru, Sam Shleifer, Punit~Singh Koura, Vishrav Chaudhary, Brian O'Horo, Jeff Wang, Luke Zettlemoyer, Zornitsa Kozareva, Mona Diab, Veselin Stoyanov, and Xian Li. 2022.
\newblock \href {http://arxiv.org/abs/2112.10668} {Few-shot learning with multilingual language models}.

\bibitem[{McCoy et~al.(2019)McCoy, Pavlick, and Linzen}]{mccoy-etal-2019-right}
Tom McCoy, Ellie Pavlick, and Tal Linzen. 2019.
\newblock \href {https://doi.org/10.18653/v1/P19-1334} {Right for the wrong reasons: Diagnosing syntactic heuristics in natural language inference}.
\newblock In \emph{Proceedings of the 57th Annual Meeting of the Association for Computational Linguistics}, pages 3428--3448, Florence, Italy. Association for Computational Linguistics.

\bibitem[{Nie et~al.(2020)Nie, Williams, Dinan, Bansal, Weston, and Kiela}]{nie-etal-2020-adversarial}
Yixin Nie, Adina Williams, Emily Dinan, Mohit Bansal, Jason Weston, and Douwe Kiela. 2020.
\newblock \href {https://doi.org/10.18653/v1/2020.acl-main.441} {Adversarial {NLI}: A new benchmark for natural language understanding}.
\newblock In \emph{Proceedings of the 58th Annual Meeting of the Association for Computational Linguistics}, pages 4885--4901, Online. Association for Computational Linguistics.

\bibitem[{Otegi et~al.(2020)Otegi, Agirre, Campos, Soroa, and Agirre}]{ixambert}
Arantxa Otegi, Aitor Agirre, Jon~Ander Campos, Aitor Soroa, and Eneko Agirre. 2020.
\newblock Conversational question answering in low resource scenarios: A dataset and case study for basque.
\newblock In \emph{Proceedings of The 12th Language Resources and Evaluation Conference}, pages 436--442.

\bibitem[{Poliak et~al.(2018)Poliak, Naradowsky, Haldar, Rudinger, and Van~Durme}]{poliak-etal-2018-hypothesis}
Adam Poliak, Jason Naradowsky, Aparajita Haldar, Rachel Rudinger, and Benjamin Van~Durme. 2018.
\newblock \href {https://doi.org/10.18653/v1/S18-2023} {Hypothesis only baselines in natural language inference}.
\newblock In \emph{Proceedings of the Seventh Joint Conference on Lexical and Computational Semantics}, pages 180--191, New Orleans, Louisiana. Association for Computational Linguistics.

\bibitem[{Sainz et~al.(2021)Sainz, Lopez~de Lacalle, Labaka, Barrena, and Agirre}]{relation-sainz-etal-2021-label}
Oscar Sainz, Oier Lopez~de Lacalle, Gorka Labaka, Ander Barrena, and Eneko Agirre. 2021.
\newblock \href {https://doi.org/10.18653/v1/2021.emnlp-main.92} {Label verbalization and entailment for effective zero and few-shot relation extraction}.
\newblock In \emph{Proceedings of the 2021 Conference on Empirical Methods in Natural Language Processing}, pages 1199--1212, Online and Punta Cana, Dominican Republic. Association for Computational Linguistics.

\bibitem[{Stowe et~al.(2022)Stowe, Utama, and Gurevych}]{figurative-stowe-etal-2022-impli}
Kevin Stowe, Prasetya Utama, and Iryna Gurevych. 2022.
\newblock \href {https://doi.org/10.18653/v1/2022.acl-long.369} {{IMPLI}: Investigating {NLI} models{'} performance on figurative language}.
\newblock In \emph{Proceedings of the 60th Annual Meeting of the Association for Computational Linguistics (Volume 1: Long Papers)}, pages 5375--5388, Dublin, Ireland. Association for Computational Linguistics.

\bibitem[{Tsuchiya(2018)}]{tsuchiya-2018-performance}
Masatoshi Tsuchiya. 2018.
\newblock \href {https://aclanthology.org/L18-1239} {Performance impact caused by hidden bias of training data for recognizing textual entailment}.
\newblock In \emph{Proceedings of the Eleventh International Conference on Language Resources and Evaluation ({LREC} 2018)}, Miyazaki, Japan. European Language Resources Association (ELRA).

\bibitem[{Volansky et~al.(2013)Volansky, Ordan, and Wintner}]{translationese}
Vered Volansky, Noam Ordan, and Shuly Wintner. 2013.
\newblock \href {https://doi.org/10.1093/llc/fqt031} {{On the features of translationese}}.
\newblock \emph{Digital Scholarship in the Humanities}, 30(1):98--118.

\bibitem[{Wang et~al.(2019)Wang, Pruksachatkun, Nangia, Singh, Michael, Hill, Levy, and Bowman}]{wang2020superglue}
Alex Wang, Yada Pruksachatkun, Nikita Nangia, Amanpreet Singh, Julian Michael, Felix Hill, Omer Levy, and Samuel~R. Bowman. 2019.
\newblock Super{GLUE}: A stickier benchmark for general-purpose language understanding systems.
\newblock \emph{arXiv preprint 1905.00537}.

\bibitem[{Wang et~al.(2018)Wang, Singh, Michael, Hill, Levy, and Bowman}]{wang-etal-2018-glue}
Alex Wang, Amanpreet Singh, Julian Michael, Felix Hill, Omer Levy, and Samuel Bowman. 2018.
\newblock \href {https://doi.org/10.18653/v1/W18-5446} {{GLUE}: A multi-task benchmark and analysis platform for natural language understanding}.
\newblock In \emph{Proceedings of the 2018 {EMNLP} Workshop {B}lackbox{NLP}: Analyzing and Interpreting Neural Networks for {NLP}}, pages 353--355, Brussels, Belgium. Association for Computational Linguistics.

\bibitem[{Williams et~al.(2018)Williams, Nangia, and Bowman}]{mnli-williams-etal-2018-broad}
Adina Williams, Nikita Nangia, and Samuel Bowman. 2018.
\newblock \href {https://doi.org/10.18653/v1/N18-1101} {A broad-coverage challenge corpus for sentence understanding through inference}.
\newblock In \emph{Proceedings of the 2018 Conference of the North {A}merican Chapter of the Association for Computational Linguistics: Human Language Technologies, Volume 1 (Long Papers)}, pages 1112--1122, New Orleans, Louisiana. Association for Computational Linguistics.

\bibitem[{Wolf et~al.(2020)Wolf, Debut, Sanh, Chaumond, Delangue, Moi, Cistac, Rault, Louf, Funtowicz, Davison, Shleifer, von Platen, Ma, Jernite, Plu, Xu, Scao, Gugger, Drame, Lhoest, and Rush}]{wolf-etal-2020-transformers}
Thomas Wolf, Lysandre Debut, Victor Sanh, Julien Chaumond, Clement Delangue, Anthony Moi, Pierric Cistac, Tim Rault, Rémi Louf, Morgan Funtowicz, Joe Davison, Sam Shleifer, Patrick von Platen, Clara Ma, Yacine Jernite, Julien Plu, Canwen Xu, Teven~Le Scao, Sylvain Gugger, Mariama Drame, Quentin Lhoest, and Alexander~M. Rush. 2020.
\newblock \href {https://www.aclweb.org/anthology/2020.emnlp-demos.6} {Transformers: State-of-the-art natural language processing}.
\newblock In \emph{Proceedings of the 2020 Conference on Empirical Methods in Natural Language Processing: System Demonstrations}, pages 38--45, Online. Association for Computational Linguistics.

\end{thebibliography}
\bibliographystyle{acl_natbib}

\clearpage
\appendix
\section{Most frequent words in original Basque}
Table \ref{tab:frequent-words-basque} shows the original words that have been translated to English in Table \ref{tab:frequent-words-english}.
\label{sec:app-freq-words}
\begin{table}[h]
\centering
\resizebox{\columnwidth}{!}{%
\begin{tabular}{lllllll}
\hline
                       & \multicolumn{2}{c}{\textbf{XNLIeu}} & \multicolumn{2}{c}{\textbf{XNLIeu$_{\mathrm{MT}}$}} & \multicolumn{2}{c}{\textbf{native}}     \\ \hline
                       & ez                                  & 0.58\%                                              & ez        & 0.54\% & euskaraz  & 0.41\% \\
                       & nuen                                & 0.24\%                                              & nuen      & 0.23\% & filma     & 0.24\% \\
\textbf{entailment}    & zerbait                             & 0.19\%                                              & batzuek   & 0.18\% & dezakezu  & 0.24\% \\
                       & batzuek                             & 0.18\%                                              & zerbait   & 0.16\% & pelikula  & 0.24\% \\
                       & daitezke                            & 0.17\%                                              & gustatzen & 0.13\% & munduko   & 0.24\% \\ 
 \hline                                                                        
                       & ez                                  & 1.61\%                                              & ez        & 1.65\% & ez        & 0.45\% \\
                       & inork                               & 0.24\%                                              & inork     & 0.23\% & euskaraz  & 0.34\% \\
\textbf{contradiction} & inoiz                               & 0.2\%                                               & nuen      & 0.18\% & euskara   & 0.28\% \\
                       & nuen                                & 0.18\%                                              & inoiz     & 0.16\% & nire      & 0.23\% \\
                       & nire                                & 0.16\%                                              & axola     & 0.14\% & bilboko   & 0.23\% \\
 \hline                                                                        
                       & ez                                  & 0.33\%                                              & ez        & 0.31\% & gustatzen & 0.37\% \\
                       & nire                                & 0.21\%                                              & dolar     & 0.2\%  & ez        & 0.37\% \\
\textbf{neutral}       & nuen                                & 0.19\%                                              & nire      & 0.2\%  & euskal    & 0.25\% \\
                       & batzuek                             & 0.18\%                                              & nuen      & 0.16\% & batzuetan & 0.25\% \\
                       & gustatzen                           & 0.15\%                                              & batzuek   & 0.16\% & jende     & 0.25\% \\ \hline
\end{tabular}
}
\caption{Proportion of most frequent words in Basque.}
\label{tab:frequent-words-basque}
\end{table}

Some common words (\textit{nuen, daitezke, dezakezu}) have been translated to English as \textit{auxiliary}. Auxiliaries are strictly grammatical words that do not hold semantic meaning. In Basque, verbal auxiliaries provide grammatical information about the tense, the mode and the person and number of the arguments of the action, the subject, the direct object and the indirect object.

\section{Prompts for the generative models}
\label{sec:prompts}
The prompts used for testing the generative models are shown in Table \ref{tab:prompts-basque}. They are a direct translation of the English prompts used in \cite{eval-harness}, which we show in Table \ref{tab:prompts-english} for completion purposes.

\begin{table}[h]
\centering
\resizebox{\columnwidth}{!}{
\begin{tabular}{ll}
\hline
\textbf{prompt}                          & \textbf{label} \\ \hline
[premise], ezta? Bai, [hypothesis]       & entailment \\
{[premise]}, ezta? Ez, [hypothesis]      & contradiction \\
{[premise]}, ezta? Gainera, [hypothesis] & neutral \\
\hline       
\end{tabular}
}
\caption{Basque prompts used in the generative models.}
\label{tab:prompts-basque}
\end{table}

\begin{table}[h]
\centering
\resizebox{\columnwidth}{!}{
\begin{tabular}{ll}
\hline
\textbf{prompt}                          & \textbf{label} \\ \hline
[premise], right? Yes, [hypothesis]      & entailment \\
{[premise]}, right? No, [hypothesis]     & contradiction \\
{[premise]}, right? Also, [hypothesis]   & neutral \\
\hline
\end{tabular}
}
\caption{English prompts in English for XNLI.}
\label{tab:prompts-english}
\end{table} 

\section{Native dataset guidelines for annotators}
\label{sec:guidelines}
\paragraph{Translation to English:}\begin{em}
The NLI (Natural Language Inference) task consists on classifying pairs of sentences according to their logical and semantical relation. The three possible relations are “Entailment” (when a sentence entails the other one), “Contradiction” (when both sentences contradict each other) and “Neutral” (when both sentences can either be true at the same time or not).

We are trying to create a dataset for this task in Basque, and we need your help.

Your work consists on reading the sentences in the “Premise” column and writing three other sentences related to the first one. Only taking into account the first sentence and your own world knowledge, you should:

\begin{itemize}
    \item Write an entailment of the premise (a sentence that is true when the premise is true) in the “Entailment” column”.
    \item Write a neutral statement in relation to the premise (a sentence whose truthfulness cannot be decided based on the premise) in the “Neutral” column.
    \item Write a contradiction of the premise (a sentence that is false when the premise is true) in the “Contradiction” column.
\end{itemize}

If there is a problem with the premise, the row can be left blank, and the box in the “Problem” column must be checked.

We would like for the sentences to have some creativity, so we discourage the use of artifacts (for example, creating contradictions by simply adding “no” to the premise).

\textbf{Example 1}

 Premise: The body language and the eyes were enough to communicate.
          \begin{itemize}
              \item Entailment: Using body language and the eyes, they were able to communicate.
              \item Neutral: We human beings are able to communicate a lot of ways.
              \item Contradiction: To understand each other they had to talk.
          \end{itemize}

\textbf{Example 2}

Premise: Solte is one of those groups that sweat from minute one in their live performances.
           \begin{itemize}
               \item Entailment: The group Solte are very lively in their concerts.
               \item Neutral: The group Solte gives a lot of concerts.
               \item Contradiction: Calmness is the main thing of the live performances of the Solte group.
           \end{itemize}
\end{em}
\paragraph{Original Basque:}
NLI (Natural Language Inference) ataza esaldi pareak sailkatzean datza, haien arteko erlazio lojiko eta semantikoan oinarrituta. Hiru erlazio aurreikusten dira esaldien artean: "Entailment" (esaldi batek bestea ondorioztatzen du), "Contradiction" (esaldiak kontraesankorrak dira) eta "Neutral" (esaldiek ez dute erlazio lojiko zuzenik).

Guk euskarazko NLI datu multzoa sortu nahi dugu, eta horretarako zure laguntza behar dugu.

Lan hau garatzeko "Premisa" zutabean dagoen esaldia irakurri behar da, eta esaldi horrekin erlazionatuta dauden beste hiru esaldi idatzi. Premisa esaldian bakarrik oinarrituz, eta zure munduko ezagutza kontuan izanik, gain zera egin behar duzu:

\begin{enumerate}
    \item Idatzi premisaren ondorio bat (premisa egia denean egia den esaldi bat) "Entailment" zutabean.
    \item "Neutroa" zutabean premisari buruzko esaldi neutro bat idatzi (premisa egia denean egia denik edo ez jakin ezin den esaldi bat).
    \item “Contradiction" zutabean premisaren kontraesana idatzi (premisa egia denean faltsua den esaldi bat).
\end{enumerate}

Erakutsitako premisarekin arazoren bat badago, lerroa hutsik utzi eta "problema" zutabeko laukian klik egin dezakezu.

Esaldi orijinalak nahi ditugu, sormena erakusten dutenak, beraz saiatu eskema berdinak ez erabiltzen (adibidez, kontraesanak sortzeko premisari "ez" hitza gehitzea).

\textbf{Adibide 1}

Premisa: Mimika eta begiak nahiko ziren komunikatzeko.

\begin{itemize}
    \item Entailment: Mimika eta begiak erabiliz, komunikatzeko gai ziren.
    \item Neutral:  Gizakiok hainbat komunikatzeko modu erabiltzeko gai gara.
    \item Contradiction:  Elkar ulertzeko hitz egin behar zuten.
\end{itemize}

\textbf{Adibide 2}

Premisa: Zuzenekoetan izerdia lehen minututik botatzen duen talde horietakoa da Solte.

\begin{itemize}
    \item Entailment: Solte taldekoak oso mugituak dira bere kontzertuetan.
    \item Neutral:  Solte taldeak kontzertu asko ematen ditu.
    \item Contradiction:  Lasaitasuna da nagusi Solte taldearen zuzenekoetan.
\end{itemize}

\end{document}